\def\year{2022}\relax
\documentclass[letterpaper]{article} 
\usepackage{aaai22}  
\usepackage{times}  
\usepackage{helvet}  
\usepackage{courier}  
\usepackage[hyphens]{url}  
\usepackage{graphicx} 
\usepackage{natbib}  
\usepackage{caption} 
\DeclareCaptionStyle{ruled}{labelfont=normalfont,labelsep=colon,strut=off} 
\usepackage{algorithm}
\usepackage{algorithmic}
\usepackage{color}
\newcommand{\eg}{\textit{e}.\textit{g}.}
\newcommand{\etal}{\textit{et al}.}
\newcommand{\ie}{\textit{i}.\textit{e}.}
\usepackage{graphics}
\usepackage{amsmath}
\usepackage{amssymb}
\usepackage{threeparttable}
\usepackage{multirow}
\usepackage{longtable}
\usepackage{rotating}
\usepackage{tabularx}
\usepackage{pifont}
\usepackage{bbding}
\usepackage{booktabs}
\usepackage{color}
\usepackage{enumitem}
\usepackage{bbm}
\usepackage{bm}
\usepackage{graphics}

\usepackage[switch]{lineno}

\usepackage{newfloat}
\usepackage{listings}
\floatstyle{ruled}
\newfloat{listing}{tb}{lst}{}
\floatname{listing}{Listing}
\title{One More Check: Making ``Fake Background'' Be Tracked Again}

\author{
    Chao Liang\textsuperscript{1~\footnote{Equally Contribute. \quad $\dagger$ Corresponding Author. }},
    Zhipeng Zhang\textsuperscript{2 \tiny{*}},
    Xue Zhou\textsuperscript{1, 3$\dagger$},
    Bing Li\textsuperscript{\rm 2},
    Weiming Hu\textsuperscript{\rm 2}}
\affiliations{
    \textsuperscript{\rm 1}School of Automation Engineering, University of Electronic Science and Technology of China (UESTC)\\
    \textsuperscript{\rm 2} NLPR, Institute of Automation, Chinese Academy of Sciences (CASIA)\\
    \textsuperscript{\rm 3} Shenzhen Institute for Advanced Study, University of Electronic Science and Technology of China (UESTC)\\

    chaoliang1996@gmail.com, zhangzhipeng2017@ia.ac.cn, zhouxue@uestc.edu.cn
}

\usepackage{bibentry}

\begin{document}

\maketitle

\begin{abstract}
The one-shot multi-object tracking, which integrates object detection and ID embedding extraction into a unified network, has achieved groundbreaking results in recent years. However, current one-shot trackers solely rely on single-frame detections to predict candidate bounding boxes, which may be unreliable when facing disastrous visual degradation, e.g., motion blur, occlusions. Once a target bounding box is mistakenly classified as background by the detector, the temporal consistency of its corresponding tracklet will be no longer maintained. In this paper, we set out to restore the bounding boxes misclassified as ``fake background'' by proposing a re-check network. The re-check network innovatively expands the role of ID embedding from data association to motion forecasting by effectively propagating previous tracklets to the current frame with a small overhead. Note that the propagation results are yielded by an independent and efficient embedding search, preventing the model from over-relying on detection results. Eventually, it helps to reload the ``fake background'' and repair the broken tracklets. Building on a strong baseline CSTrack, we construct a new one-shot tracker and achieve favorable gains by $70.7 \rightarrow 76.4$, $70.6 \rightarrow 76.3$ MOTA on MOT16 and MOT17, respectively. It also reaches a new state-of-the-art MOTA and IDF1 performance. Code is released at https://github.com/JudasDie/SOTS.
\end{abstract}



\section{Introduction}
\noindent Multi-object tracking (MOT), aiming to estimate the trajectory of each target in a video sequence, is one of the most fundamental yet challenging tasks in computer vision~\cite{luo2020multiple}. The related technique underpins significant applications from video surveillance to autonomous driving.

\begin{figure}[t]
\begin{center}
\includegraphics[width=1\linewidth]{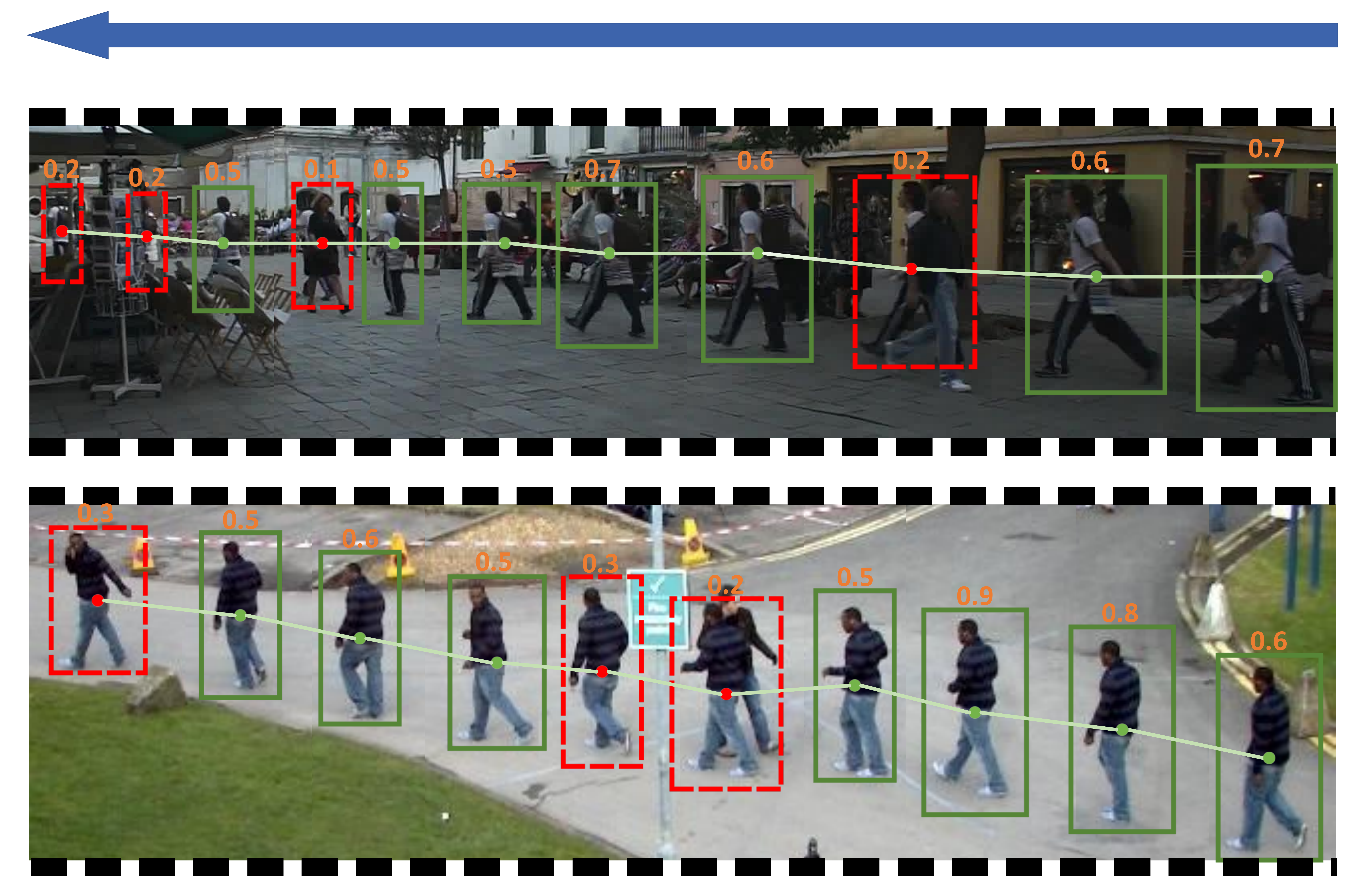}
\end{center}
\vspace{-0.4cm}
\caption{Illustration of tracklets generated from CSTrack on two clips. Wherein, ``fake background'' is represented by {\color{red} red box}. The blue arrow indicates the motion direction of the target. Best viewed in color and zoom in.}
\label{fig:tracklet_crack}
\vspace{-0.4cm}
\end{figure}

The current MOT methods are categorized into two-step and one-shot frameworks. The two-step framework~\cite{bewley2016simple,wojke2017simple,yu2016poi,tang2017multiple,xu2019spatial}, following the tracking-by-detection paradigm (or more precisely, tracking-after-detection), disentangles MOT into candidate boxes prediction and tracklet association. Though favored in astonishing performance, they suffer from massive computation cost brought by separately extracting ID (identity) embedding of each candidate box through an isolated ReID (Re-identification) network~\cite{wojke2017simple,zheng2017person}. Recently, the one-shot methods~\cite{xiao2017joint,wang2019towards,zhang2020fairmot,liang2020rethinking}, which integrate detection and ID embedding extraction into a unified network, have drawn great attention because of their balanced speed and accuracy. By sharing features and conducting multi-task learning, they are capable of running at quasi real-time speed. We observed that most existing one-shot trackers work under a strong complete detection assumption, in other words, all targets are presumed to be correctly localized by the detector. However, various real-world challenges may break such assumption and cause these approaches to fail. Fig.~\ref{fig:tracklet_crack} shows typical failure cases of the one-shot trackers (\eg, CSTrack~\cite{liang2020rethinking}), where targets (\textcolor{red}{red boxes}) are considered as the background in some frames due to small foreground probabilities. The missed targets will break temporal consistency of a tracklet. 

Revisiting the failures of one-shot trackers, we find that the integrated detector solely considers single-frame visual cues. Nevertheless, the challenging scenes in practical tracking, \eg, occlusion, motion blur, background clutter, will cause visual feature degradation, which may eventually mislead the detector to classify the targets as background. Hence, heavily relying on single frame detections is not reliable. In contrast, human vision has a dynamic view of targets, not only taking current visual cues into consideration, but also being continuously aware of the temporal consistency of moving targets. This inspires us that exploring temporal cues might be a potential solution for reloading the misclassified targets of the detector. 

In this work, motivated by inheriting the merits of one-shot models and mining temporal cues to make missed targets be tracked again, we propose a double-check mechanism to construct a new one-shot tracker. As an auxiliary to initial detections, a re-check network is delicately designed to learn the transduction of previous tracklets to the current frame. Given a target that appeared in previous frames, the propagation results re-check the entire scene in the current frame to make a candidate box prediction. If a groundtruth target box does not exist in the first-check predictions (\emph{i.e.,} the results of object detector), as a potential misclassified target, it has a chance to be restored. Technically tracklet propagation is achieved by ID embedding search across frames, which is inspired by cross-correlation operation from Siamese trackers~\cite{SiamFC, SiamDW, Ocean}. Some previous MOT methods~\cite{chu2019famnet,yin2020unified} attempt to introduce a separate Siamese network to learn additional clues of all targets for motion search, which are tedious and complex. Differently, we innovatively expand the role of ID embeddings from data association to motion forecasting. By reusing ID embeddings for propagation, the overhead of modeling temporal cues is minimized.  Even with multiple tracklets, our re-check network can still propagate with one forward pass by a simple matrix multiplication.


Finally, we propose our new one-shot tracker, namely \textbf{OMC} (the initials of \textbf{O}ne \textbf{M}ore \textbf{C}heck), which is built on a baseline model CSTrack~\cite{liang2020rethinking}. It's worth noted that our proposed tracker efficiently integrates detection, embedding extraction and temporal cues mining into a unified framework. We evaluate the proposed OMC on three MOT Challenge~\footnote{\url{https://motchallenge.net}} benchmarks: MOT16~\cite{milan2016mot16}, MOT17~\cite{milan2016mot16} and MOT20~\cite{dendorfer2020mot20}. Our method achieves new state-of-the-art MOTA and IDF1 on all three benchmarks. Furthermore, compared with other trackers using temporal cues under the same public detection protocol~\cite{milan2016mot16}, our method still achieves better tracking performance on MOTA and IDF1.

The main contributions of our work are as follows:
\begin{itemize}
\item We propose a simple yet effective double-check mechanism to restore the misclassified targets induced by the imperfect detection in MOT task. Our proposed re-check network flexibly expands ID embeddings from data association to motion forecasting, propagating previous tracklets to the current frame with a small overhead.

\item Our proposed re-check network is a ``plug-and-play'' module that can work well with other one-shot trackers. We build it on a strong baseline CSTrack and construct a new one-shot tracker. The experimental results demonstrate that our tracker \textbf{OMC} not only outperforms  CSTrack largely, but also achieves new state-of-the-art MOTA and IDF1 scores on all three benchmarks.
\end{itemize}

\section{Related Work}
\subsection{Detection-based Tracking}
Recent MOT trackers can be summarized into two streams, \ie, two-step and one-shot structures. The former one follows the tracking-by-detection paradigm, where object bounding boxes are first predicted by a detector and then linked into tracklets by an association network ~\cite{bewley2016simple,wojke2017simple,yu2016poi,tang2017multiple,xu2019spatial}. These methods mainly focus on improving association accuracy. Though favored in good tracking performance, they suffer from computation cost brought by extracting ID embeddings for all bounding boxes with an additional ReID network~\cite{wojke2017simple,zheng2017person}. Alternatively, the one-shot paradigm which integrates detection and ID embedding extraction into a unified network, is a new trend in MOT~\cite{xiao2017joint, wang2019towards,zhang2020fairmot,liang2020rethinking}. Tong \etal ~\cite{xiao2017joint} first propose an end-to-end framework to jointly handle detection and ReID tasks. By adding extra fully connected layers to a two-stage detector (Faster RCNN~\cite{ren2016faster}), the model can simultaneously generate detection boxes and the corresponding ID embeddings. Recent proposed JDE~\cite{wang2019towards} converts the one-stage detector YOLOv3~\cite{redmon2018yolov3} to a one-shot tracker by redesigning the prediction head. The follow-up CSTrack~\cite{liang2020rethinking} further eases the competition between detection and ID embeddings learning by applying a cross-attention network to JDE~\cite{wang2019towards}. However, detection-based methods assume that all the targets can be precisely localized by the detector, which is not valid in practical tracking. When challenging scenes degrade the visual cues, the detector may miss some targets. In this work, we exploit cross-frame temporal cues to alleviate this issue. Below, we briefly review other methods that utilize temporal features to improve MOT trackers and discuss the differences between us.


\begin{figure*}[t]
\begin{center}
\includegraphics[width=1\linewidth]{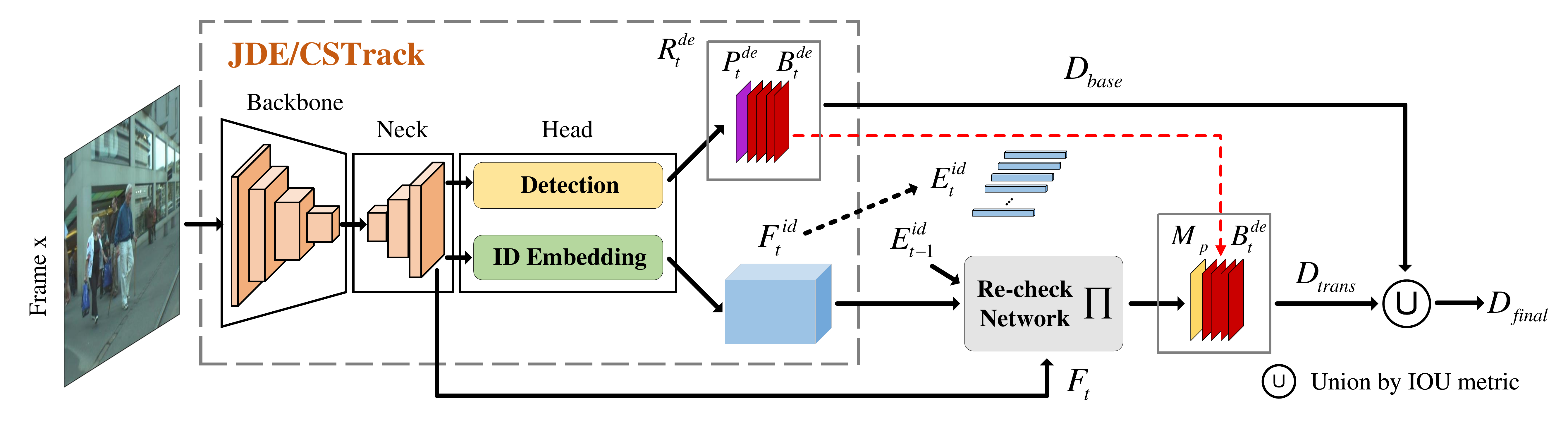}
\end{center}
\vspace{-0.4cm}
\caption{Overview of the proposed OMC. It consists of the baseline CStrack tracker and a re-check network. The CStrack tracker first generates detection result $\bm{R}_{t}^{de}$ and candidate embeddings $\bm{F}_{t}^{id}$. Then, re-check network improves temporal consistency by reloading the misclassified targets induced by the detector.}
\label{fig:OMC}
\vspace{-0.1cm}
\end{figure*}


\subsection{Temporal Cues Mining}
Some previous works attempt to utilize extra information, \eg, motion~\cite{chen2018real,hornakova2020lifted}, temporal visual features~\cite{chu2019famnet}, to improve detection performance in MOT task. Early works~\cite{dehghan2015target,ristani2018features,chen2018real,hornakova2020lifted} consider MOT as a global optimization problem and obtains auxiliary candidates generated by Kalman filter, spatial interpolation, or visual extrapolation. Albeit efficient and straightforward, these methods leverage both past and future frames for batch processing that is not suitable for causal applications. For the aim of online tracking, recent works apply off-the-shelf SOT trackers, \eg, SiamFC~\cite{SiamFC}, to estimate target motion. In the literature, there are two major branches of inserting SOT trackers into the MOT system. The first one~\cite{chu2019famnet,yin2020unified,dasot} aims to modify SOT networks and integrate them into an isolated association network for joint learning. In this regard, it's essential to equip an extra affinity learning model for handling drift. The other one~\cite{chu2017online,sadeghian2017tracking,zhu2018online,automatch} exploits separate SOT trackers to create auxiliary clues for handling complex MOT scenes. Despite the performance gains, they are not suitable for real-time applications because of the massive computation brought by applying a SOT network to learn auxiliary clues for all targets. In our work, instead of assigning an extra and complex SOT network, we expand the role of ID embeddings from data association to motion forecasting by similarity matching. It makes our tracker capable of tracking multiple targets with only a simple forward pass.


\section{Methodology}
\label{sec:OMC}
In this section, we describe the proposed tracking framework, as illustrated in Fig.~\ref{fig:OMC}. 


\subsection{Overview}
\label{sec:overview}
The proposed model is conducted on a recent MOT tracker, namely CSTrack~\cite{liang2020rethinking}, which is a variant of the recent JDE framework~\cite{wang2019towards}. In this section, we firstly describe the reasoning procedure of JDE and CSTrack. Then we elaborate on the details of integrating our proposed model into the baseline tracker. 

\vspace{1mm}
\noindent\textbf{Baseline Tracker.} JDE~\cite{wang2019towards} devotes effort to building a real-time one-shot MOT framework by allowing object detection and ID embedding extraction to be learned in a shared model, as shown in Fig.~\ref{fig:OMC}. Given a frame $\bm{x}$, it is firstly processed by a feature extractor $\Psi$ (\emph{e.g.,} \emph{Backbone} and \emph{Neck}), which generates the feature $\bm{F}_t$,
\begin{equation}
\bm{F}_t=\Psi(\bm{x}).
\end{equation}
Then $\bm{F}_t$ is fed into the \emph{Head} network $\Phi$ to simultaneously predict detection results and ID embeddings,
\begin{gather}
    [\bm{R}_{t}^{de}, \bm{F}_{t}^{id}]=\Phi(\bm{F}_t),
\end{gather}
where $\bm{R}_{t}^{de}$ is the detection results (including one map $\bm{P}_{t}^{de} \in \mathbb{R}^{H \times W \times 1}$ for foreground probabilities and the others $\bm{B}_{t}^{de} \in \mathbb{R}^{H \times W \times 4}$ for raw boxes). $\bm{F}_{t}^{id} \in \mathbb{R}^{H \times W \times C}$ (C=512) denotes ID embeddings. The detection results $\bm{R}_{t}^{de}$ are processed by greedy-NMS~\cite{ren2016faster} to generate the \textbf{basic detections} $\bm{D}_{base}$. Each box in $\bm{D}_{base}$ corresponds to a $1\times 1 \times C$ embedding in $\bm{F}_{t}^{id}$. We denote $\bm{E}_{t}^{id}$ as a set that contains embeddings of all boxes in $\bm{D}_{base}$. Finally, the boxes $\bm{D}_{base}$ and the ID embeddings $\bm{E}_{t}^{id}$ are utilized to associate with the prior tracklets by greedy bipartite matching. The recent CSTrack~\cite{liang2020rethinking} introduces cross-attention to ease the competition between detection and ReID, which significantly improves the JDE with small overhead. Here, we use CSTrack as our baseline tracker. 


\begin{figure*}[t]
\begin{center}
\includegraphics[width=1\linewidth]{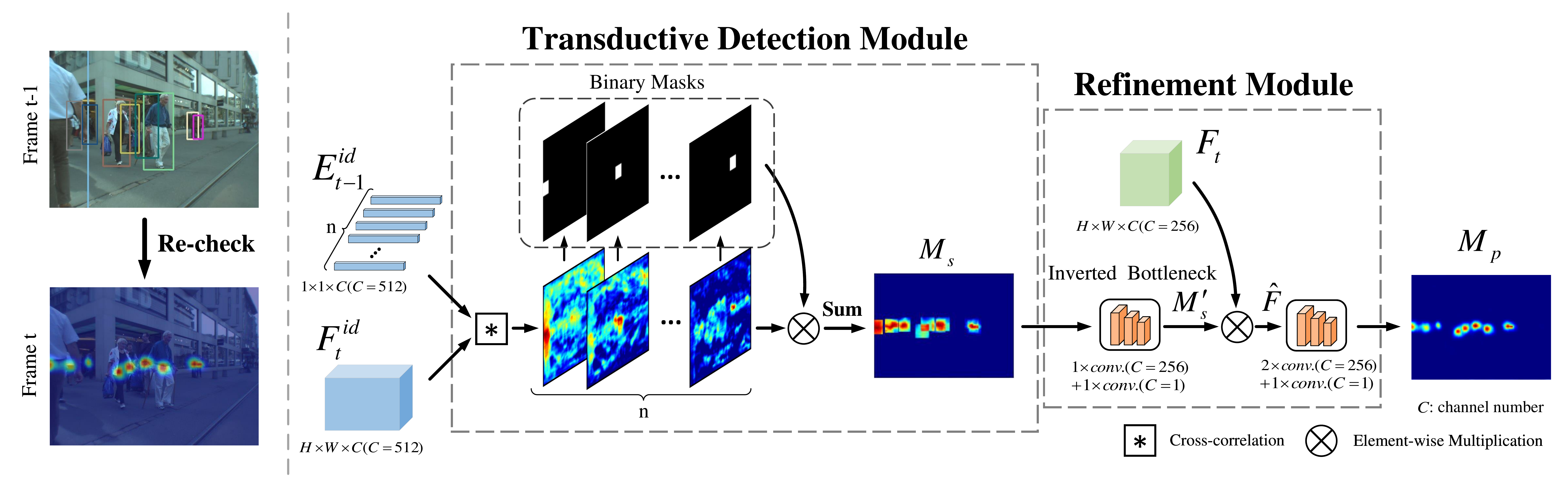}
\end{center}
\vspace{-0.4cm}
\caption{\textbf{The architecture of the proposed re-check network.}  Re-check network consists of two major components: the transductive detection module and the refinement module. More details are described in section ``Methodology''.}
\label{fig:recheck}
\end{figure*}


\vspace{1mm}
\noindent\textbf{OMC.} In this work, we propose a re-check network to repair the ``fake background'' induced by the detector in JDE and CSTrack. As shown in Fig.~\ref{fig:OMC}, we reuse the ID embeddings from previous targets ($\bm{E}_{t-1}^{id}$) as temporal cues. The re-check network $\Pi$ transfers the prior tracklets by measuring the similarity between $\bm{E}_{t-1}^{id}$ and $\bm{F}_{t}^{id}$. Specifically, we modify the cross-correlation layer, that is used by Siamese method in single object tracking~\cite{SiamFC}, to make it capable of tracking multiple targets in a single forward pass. We experimentally observed that if one target disappears in the current frame, it tends to introduce a false-positive response in the similarity map. To alleviate this issue, we fuse the visual feature $\bm{F}_{t}$ with the similarity map, and then refine them to a finer guidance map. For simplicity, we omit the operation of the re-check network as,
\begin{equation}
\bm{M}_{p}=\Pi\left(\bm{F}_{t}^{i d}, \bm{E}_{t-1}^{i d}, \bm{F}_{t}\right),
\label{eq:recheck_ow}
\end{equation}
where the final prediction $\bm{M}_{p}$ represents the transduction of prior tracklets to the current frame. We consider $\bm{M}_{p}$ as the foreground probability, and send it to greedy-NMS with original bounding boxes $\bm{B}_{t}^{de}$ (\textcolor{red}{red response maps}). The outputs of NMS, namely \textbf{transductive detections} $\bm{D}_{trans}$, are combined with basic detections $\bm{D}_{base}$ through the proposed IOU vote mechanism to generate the final candidate bounding boxes $\bm{D}_{final}$. $\bm{D}_{final}$ and the corresponding ID embeddings extracted from $\bm{F}_{t}^{i d}$ are used for latter association. When the basic detections mistakenly classify the targets as background, the transductive detections can recheck the ``fake background'' and restore the missed boxes.

\subsection{Re-check Network}
\label{sec:recheck}
To improve the temporal consistency broken by ``fake background'', we propose a lightweight re-check network to restore the missed targets induced by the detector. Precisely, re-check network consists of two modules, \ie, the transductive detection module for tracklets propagation and the refinement module for false positives filtering. 


\subsubsection{Transductive Detection Module}
\label{sec:tdn}

The transductive detection module aims to propagate previous tracklets to current frame, in other words, predict locations of existing targets. Concretely, target locations are predicted by measuring the similarities between previous tracklet embeddings $\bm{E}_{t-1}^{id}=\left\{\bm{e}_{t-1}^{1}, \cdots, \bm{e}_{t-1}^{n} \right\}$ and current candidate embeddings $\bm{F}_{t}^{id}$, where $n$ indicates the number of previous tracklets (include all active tracklets in the inference stage, not just the last frame). We get a location response map $m_i$ for each target through a cross-correlation operator $*$,
\begin{equation}
\bm{m}_{i}=(\bm{e}_{t-1}^{i} * \bm{F}_{t}^{id})|_{i=1}^n.
\label{eq:simi}
\end{equation}
Wherein, location with the maximum value in $\bm{m}_{i}$ indicates the predicted state of a previous tracklet. Eq.~\ref{eq:simi} yields a set of similarity maps $\bm{M}=\left\{\bm{m}_{1}, \cdots, \bm{m}_{n} \right\}$, in which each denotes the transductive detection result of a previous tracklet. Notably, the modified cross-correlation in our model can be implemented with a simple matrix multiplication. We attach the PyTorch codes in Alg.~\ref{al:cross-correlation}.

\begin{algorithm}[t]
\caption{Modified Cross-correlation, PyTorch-like}
\label{al:cross-correlation}
\begin{algorithmic}
\small{
\STATE \textcolor[rgb]{0.9,0.15,0.8}{def} \ D ($\bm{E}_{t-1}^{id}$, $\bm{F}_{t}^{id}$): 
\STATE \qquad E = torch.Tensor($\bm{E}_{t-1}^{id}$).view(n, c) \textcolor[rgb]{0.4,0.7,0.7}{\ \ \#\ \ convert list to matrix}
\STATE \qquad F = $\bm{F}_{t}^{id}$.view(c, h $*$ w) \textcolor[rgb]{0.4,0.7,0.7}{\ \ \#\ \ reshape tensor to matrix}
\STATE \qquad M = torch.matmul(E, F) \textcolor[rgb]{0.4,0.7,0.7}{\ \ \#\ \ matrix multiplication}
\STATE \qquad \textcolor[rgb]{0.9,0.15,0.8}{return} M.view(h, w, n) \textcolor[rgb]{0.4,0.7,0.7}{\ \ \#\ \ reshape matrix to tensor}
}
\end{algorithmic}
\end{algorithm}
We then discretize $\bm{m}_i$ to a binary mask $\hat{\bm{m}}_{i}$ by shrinking the scope of high responses. The underlying reason for this operation is that objects with similar appearance may bring high response. Thus, shrinking the scope of high responses can reduce ambiguous predictions. More formally, the binary mask $\hat{\bm{m}}_{i}$ is obtained by,
\begin{equation}
\hat{\bm{m}}_{i}^{xy}=\left\{\begin{array}{cl}1 & \text { if } \left\|x-c_{x}\right\| \leq r,\left\|y-c_{y}\right\| \leq r \\
0 &{\text{otherwise }}\\
\end{array}\right.
\label{eq:shrink}
\end{equation}
where $\hat{\bm{m}}_{i}^{xy}$ denotes the value at $(x,y)$ of $\hat{\bm{m}}_{i}$, and $c_{x}, c_{y}$ indicate the locations of maximum value in $\bm{m}_i$. $r$ is the shrinking radius. The region within and outside the square is set to 1 and 0, respectively. Afterwards, we multiply the binary mask $\hat{\bm{m}}_{i}$ to the original similarity map $\bm{m}_{i}$ to reduce ambiguous responses. Finally, we aggregate the response maps by element-wise summation along channel dimension, 
\begin{equation}
\bm{M}_s = \sum_{i=1}^{n} (\hat{\bm{m}}_{i} \cdot \bm{m}_{i}).
\label{eq:shrink_ms}
\end{equation}
The aggregated similarity map $\bm{M}_{s}$ reveals the probability of a location in the current frame that contains a bounding box associated with previous tracklets. 



\subsubsection{Refinement Module} 
\label{sec:refinement}
We observed that objects disappearing in the current frame tend to bring false positives during tracklet transduction. To alleviate this issue,  we arrange the refinement module to introduce the original visual feature $\bm{F}_t \in \mathbb{R}^{H \times W \times C}$ (C=256) to provide informative semantics for finer localization. We firstly encode the similarity map $\bm{M}_s$ with an inverted bottleneck module~\cite{sandler2018mobilenetv2}. Concretely, a $3 \times 3$ convolution layer maps $\bm{M}_s$ to high dimensional space, \ie, the channels of 256. Then another $3 \times 3$ convolution layer follows to down-sample the channel to 1, as $\bm{M}_s^{'} \in \mathbb{R}^{H \times W \times 1}$. The refined similarity map $\bm{M}_s^{'}$ is multiplied by the visual feature $\bm{F}_t$ to get the enhanced feature $\hat{\bm{F}} \in \mathbb{R}^{H \times W \times C}$ (C=256),
\begin{equation}
\hat{\bm{F}}=\bm{F}_{t} \cdot \bm{M}_s^{'}.
\label{eq:ftenhance}
\end{equation}
Later, the enhanced feature $\hat{\bm{F}}$ passes through several convolution layers to obtain the final prediction $\bm{M}_{p}$.

\subsubsection{Optimization}
Besides the loss for the baseline tracker CSTrack~\cite{liang2020rethinking}, we introduce a supervised function to train the re-check network. The ground-truth for the similarity map $\bm{M}_p$ is defined as a combination of multiple Gaussian distributions. Specifically, for each target, its supervised signal is a Gaussian-like mask,
\begin{equation}
    \bm{t}_i= \exp \left(-\frac{\left(x-c_i^{x}\right)^{2}+\left(y-c_i^{y}\right)^{2}}{2 \sigma_i^{2}}\right)
    \label{eq:s-gt}
\end{equation}
where $c_{i} = ({c_i^{x}},{c_i^{y}})$ denotes the center location of a target and $\sigma_i$ is the object size-adaptive standard deviation~\cite{law2018cornernet}. Eq.~\ref{eq:s-gt} generates a set of ground-truth masks $\bm{t}=\{\bm{t}_1, ..., \bm{t}_n\}$. Then we sum all elements in $\bm{t}$ along channel dimension to get the supervised signal $\bm{T}$ for $\bm{M}_p$. To reduce the overlap between two Gaussian distributions, we set an upper limit of $\sigma_{i}$ to 1. We employ the Logistic$-$MSE Loss~\cite{allen1971mean} to train re-check network,
\begin{equation}
\mathcal{L}_{g}=-\frac{1}{n} \sum_{x y}\left\{\begin{array}{cl}
\left(1-\bm{M}_p^{xy}\right) \log \left(\bm{M}_p^{xy}\right), \hspace{0.6em}  \text { if } \bm{T}^{xy}=1 \\
\left(1-\bm{T}^{xy}\right)\bm{M}_{p}^{xy}
\log \left(1-\bm{M}_{p}^{xy}\right),  \text {else}  
\end{array}\right.
\end{equation}
where $\bm{M}^{xy}$ and $\bm{T}^{xy}$ indicate the value of a location in $\bm{M}_p$ and $\bm{T}$, respectively.

\subsection{Fusing Basic and Transductive Detections}
\label{sec:fuse}
In this section, we detail how to fuse the transductive detections $\bm{D}_{trans}$ and basic detections $\bm{D}_{base}$ to get the final candidate boxes $\bm{D}_{final}$ for association. We firstly calculate the targetness score $s$ for each bounding box $\bm{b}_i$ in $\bm{D}_{trans}$ by IOU metric, as
\begin{equation}
s = 1 - \max \left( IOU\left(\bm{b}_i, \bm{D}_{base}\right)\right),
\end{equation}
where a higher $s$ indicates the box $\bm{b}_i$ does not appear in the basic detections, which is a probable missed bounding box. Then, the boxes with a score above threshold $\epsilon$ are retrained as complement of basic detections. We set $\epsilon$ to 0.5. When the basic detections miss some targets, the transductive detections can restore them to keep the temporal consistency of tracklets. 





\subsection{Comparisons with Related Works}
In this section, we further discuss the differences with other works which share similar spirit with our method.

\noindent\textbf{OMC \emph{vs.} UMA~\cite{yin2020unified} and DASOT~\cite{dasot}.} Recent works UMA and DASOT also adopt SOT trackers or mechanism in MOT tracking. However, our method differs from them in two fundamental ways. 1) \emph{Local or Global search.} UMA and DASOT only consider a small neighborhood region when searching a target. However, local search is not effective when fast motion happens. Conversely, in our work, the tracklet transduction is accomplished with global search, which is more robust for fast motion cases. 2) \emph{Unified or Separated Framework.} UMA and DASOT only integrate temporal cues mining and data association into a model, which is separated with the object detector. This is obviously tedious and time-consuming since the raw image input needs to be performed forward inference two or even more times. Differently, in our work, detection-transduction-association are unified in a tracking framework, which obtains all outputs with only one single pass and enjoys easier implementation.



\noindent\textbf{OMC \emph{vs.} Tracktor~\cite{bergmann2019tracking}.} Both OMC and Tracktor attempt to propagate previous tracklets to the current frame in a simple one-shot framework. Tracktor considers the bounding boxes in the last frame as regions of interest (ROIs) in the current frame, and then extracts features inside the ROIs. The locations of existing targets are predicted by directly regressing the ROI features. However, the tracklet transduction of Tracktor still relies on the single-frame visual cues. Differently, OMC transfers the previous tracklets by measuring ID embedding similarities between the last frame and the current frame. By reusing the object ID embeddings of the last frame as temporal cues, OMC can restore missed targets more effectively. 

\section{Experiments}

\subsection{Implementation Details}
\label{sec:Imple}

\begin{figure}[t]
\begin{center}
\includegraphics[width=1\linewidth]{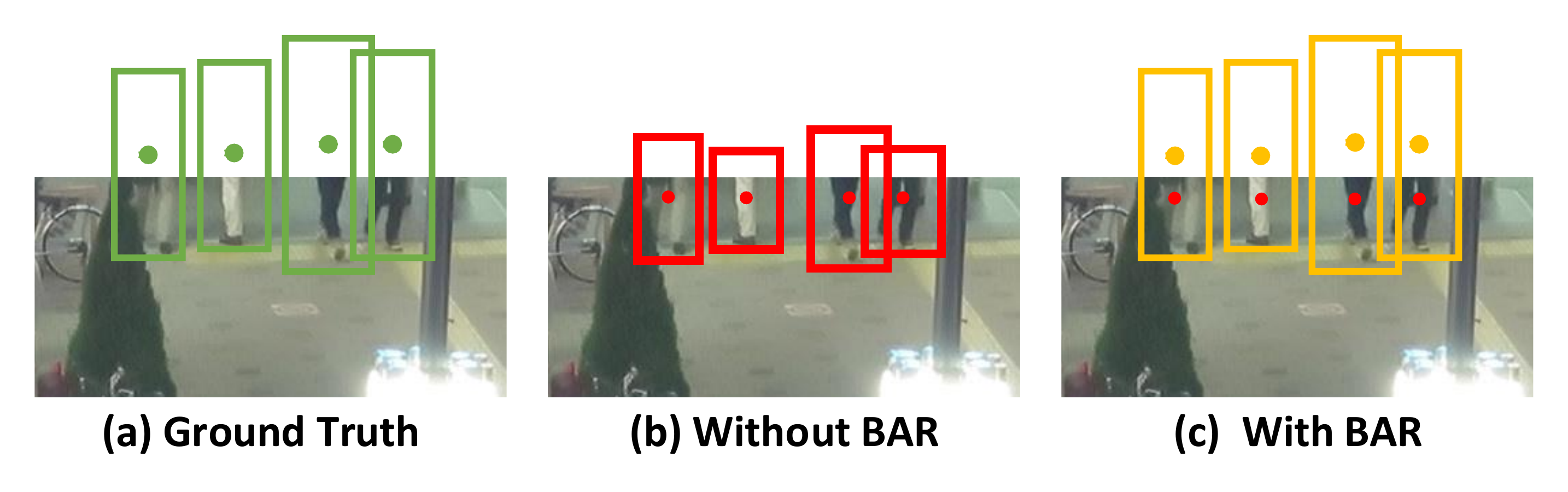}
\end{center}
\vspace{-0.2cm}
\caption{\textbf{Visualization detection results at the boundary.} (a) Ground-truth bounding boxes. (b) The incomplete bounding box prediction. (c) The bounding box prediction with boundary-aware regression (BAR). The red points represent the anchor centers and green/yellow points represent centers of the bounding boxes.}
\label{fig:Boundary}
\end{figure}

\noindent\textbf{Baseline Tracker Modification.}
In the vanilla JDE~\cite{wang2019towards} and CSTrack~\cite{liang2020rethinking}, the offset between an anchor center $\bm{a}=(a_x,a_y)$ and the center of corresponding bounding box $\bm{b}=(b_x,b_y)$ is restricted to $0\sim 1$ (on the feature map) by the sigmoid function,
\begin{equation}
\bm{\Delta} = \bm{b}-\bm{a} = \operatorname{Sigmoid}(\bm{r})
\label{eq:ori}
\end{equation}
where $\bm{r}$ indicates the network's regression output and $\bm{\Delta}=(\Delta_x,\Delta_y)$ denotes the predicted offset. However, at the boundary of an image, the offset is often larger than 1. As shown in Fig.~\ref{fig:Boundary}, the centers of groundtruth boxes (\textcolor{green}{green}) are outside the image boundary. However, due to the hard restriction of Sigmoid, the predicted boxes (\textcolor{red}{red}) hardly cover the whole objects. When an object appears with only part-body, the incomplete box prediction will be considered as false positive because of the large differences between the ground-truth bounding box and the incomplete box, which eventually degrades tracking performance. To alleviate this issue, we modify the regression mechanism to a boundary-aware regression (BAR) as,
\begin{equation}
\bm{\Delta} = \bm{b}-\bm{a} =\left(\operatorname{Sigmoid}\left(\bm{r}\right)-0.5\right) \times h,
\label{eq:bar}
\end{equation}
where $h$ is the learnable scale parameter. The scale parameter allows the network to predict offsets larger than 1. As shown in Fig.~\ref{fig:Boundary} (c), BAR is capable of predicting invisible part of the objects based on the visible part.

\label{sec:BAR}

\vspace{1mm}
\noindent\textbf{Training and Testing.} We build our tracker by integrating the proposed re-check network into CSTrack~\cite{liang2020rethinking}. For the sake of fairness, we use the same training data as CSTrack, including ETH~\cite{ess2008mobile}, CityPerson~\cite{zhang2017citypersons}, CalTech~\cite{dollar2009pedestrian}, MOT17~\cite{milan2016mot16}, CUDK-SYSU~\cite{xiao2017joint}, PRW~\cite{zheng2017person} and CrowdHuman~\cite{shao2018crowdhuman}. The training procedure consists of two stages, \emph{i.e.}, basic tracker training and re-check network optimization. In the first stage, we equip CSTrack with the Boundary-Aware Regression (basic tracker) and train it following the standard settings of CSTrack. Concretely, the network is trained with a SGD optimizer for 30 epochs. The batch size is 8. The initial learning rate is $5 \times 10^{-4}$, and it decays to $5 \times 10^{-5}$ at the $20^{th}$ epoch. In the second stage, we train the proposed re-check network while fixing the basic tracker's parameters on MOT17~\cite{milan2016mot16} training set. During training, we randomly sample image pairs from adjacent frames in the same video sequence, one for generating exemplar embeddings $\bm{E}_{t-1}^{id}$ and the other for generating candidate embeddings $\bm{F}_{t}^{id}$. Each iteration contains 8 pairs. Other training schedules follow the settings in the first training stage. We set $r$ in Eq.~\ref{eq:shrink} to 3 and initialize the scale parameter $h$ in Eq.~\ref{eq:bar} to 10. Other hyperparameters and testing stage follow settings in CSTrack without other specifications. 

Our tracker is implemented using Python 3.7 and PyTorch 1.6.0. The experiments are conducted on a single RTX 2080Ti GPU and Xeon Gold 5218 2.30GHz CPU.

\vspace{1mm}
\noindent\textbf{Evaluation Datasets and Metrics.}
We evaluate our tracker on three MOT Challenge benchmarks, \emph{i.e.,} MOT16~\cite{milan2016mot16}, MOT17~\cite{milan2016mot16} and the recent released MOT20~\cite{dendorfer2020mot20}. Following the common practices in MOT Challenge~\cite{milan2016mot16}, we employ the CLEAR metric~\cite{bernardin2008evaluating}, particularly MOTA (the primary metric of MOT) and IDF1~\cite{ristani2016performance} to evaluate the overall performance. We also report other common metrics for evaluation, which include the ratio of Most Tracked targets (MT), the ratio of Most Lost targets (ML)
, False Positives (FP), False Negatives (FN) and running speed (FPS).


\subsection{Analysis of the Proposed Method}
\label{sec:analysis}

\begin{table}[!t]
    \begin{center}
        \caption{Component-wise analysis of the proposed model.} 
        \fontsize{7.5pt}{4mm}\selectfont
        \begin{threeparttable}
            \begin{tabular}{@{}c@{} | @{}c@{}  @{}c@{} | @{}c@{} @{}c@{}  @{}c@{} @{}c@{} @{}c@{} @{}c@{} @{}c@{} @{}c@{}}
                \cline{1-11}
                \#NUM ~& ~~R~ & ~BAR~~& ~MOTA$\uparrow$ & ~IDF1$\uparrow$~  & ~~MT$\uparrow$~~ &  ~ML$\downarrow$~ &  ~FP$\downarrow$~ &  ~FN$\downarrow$~ & ~FPS$\uparrow$~ & Param
                \\
                \cline{1-11} 
                \ding{172}  & & & 70.6 & 71.6 & 37.5 & 18.7 & 24804~~&~137832 & \bf{15.8} & 74.6M \\
\ding{173}  &~~\checkmark & & 75.5 & 72.9 & 42.0 & 15.5 & 27334~~&~107284& 13.3 & 77.5M\\
\ding{174} & &\checkmark  & 73.1 & 72.4 & 39.9 & 16.4 & \bf{19772}~~&~128184 & 15.2 & 74.6M\\
\ding{175}  &~~\checkmark &\checkmark & \bf{76.3} & \bf{73.8} & \bf{44.7} & \bf{13.6} & 28894~~&~\bf{101022} & 12.8  & 77.5M\\
                \cline{1-11}

            \end{tabular}

        \end{threeparttable}
        \vspace{-1em}
        \label{tab:Cw}
    \end{center}
\end{table}

\noindent\textbf{Component-wise Analysis.} 
To verify the efficacy of the proposed method, we perform a component-wise analysis on MOT17 testing set, as presented in Tab.~\ref{tab:Cw}. When equipping the baseline tracker (\ding{172}) with the proposed re-check network (R), it significantly decreases FN from 137832 to 107284 (\ding{173} \emph{vs.} \ding{172}), which achieves favorable 4.9 points gains on MOTA and 4.5 points gains on MT. This confirms the effectiveness of the re-check network on restoring ``fake background''. The introduced boundary-aware regression (BAR) aims to reason the invisible part of objects when they appear at the boundary of image. Tab.~\ref{tab:Cw} shows that the BAR brings gains of 2.5 points on MOTA and 0.8 points on IDF1 (\ding{174} \emph{vs.} \ding{172}), respectively. Overall (\ding{172} \emph{vs}. \ding{175}), compared with the baseline tracker, our model significantly improves tracking performance, \ie, MOTA +5.7 points, IDF1 +2.2 points and MT +7.2 points, with a small overhead, \ie, 12.8 FPS \emph{vs.} 15.8 FPS and model parameters 77.5M \emph{vs.} 74.6M.

\begin{table}[!t]
    \begin{center}

        \caption{Analysis of re-check network on MOT17 testing set. \ding{172} indicates our baseline (CSTrack) with BAR. While with ``w/o'' means that the method discards this module. }
        
        \fontsize{8.2pt}{4.3mm}\selectfont
        \begin{threeparttable}
            \begin{tabular}{@{}c@{} | @{}c@{} | @{}c@{}  @{}c@{} @{}c@{}  @{}c@{} @{}c@{} @{}c@{}}
                \cline{1-8}
                \#NUM ~& ~Method~ & ~MOTA$\uparrow$~  & ~IDF1$\uparrow$~  & ~~MT$\uparrow$~~ &  ~ML$\downarrow$~ & ~FP$\downarrow$~ &  ~FN$\downarrow$~
                \\
                \cline{1-8} 
                \ding{172} & ~Baseline-BAR~ & 73.1 & 72.4 & 39.9 & 16.4 & \bf{19772}~~&~128184\\
                \ding{173} & ~+ R \emph{w/o} Global~ & 75.4  & 73.2  & 43.4  & 15.4  & 31013 & 103766\\
                \ding{174} & ~+ R \emph{w/o} Shrink~ & 73.6  & 72.6  & \bf{47.8}  & \bf{11.2}  & 48915  & \bf{95829}\\
                \ding{175} & ~+ R \emph{w/o} $\bm{F}_t$~ & 69.3  & 70.7  & 45.2  & 13.5  & 67885  & 100793\\
                \ding{176} & ~+ R \emph{w/o} IBM~ & 75.7  & 73.4  & 45.1  & 13.5  & 32005  & 100590 \\
                \ding{177} & + R   & \bf{76.3} & \bf{73.8} & 44.7 & 13.6 & 28894~~&~101022\\
                \cline{1-8}

            \end{tabular}
        \end{threeparttable}
        \label{tab:recheck_a}
    \end{center}
    \vspace{-0.2cm}
\end{table}

\begin{figure}[t]
\begin{center}
\includegraphics[width=1\linewidth]{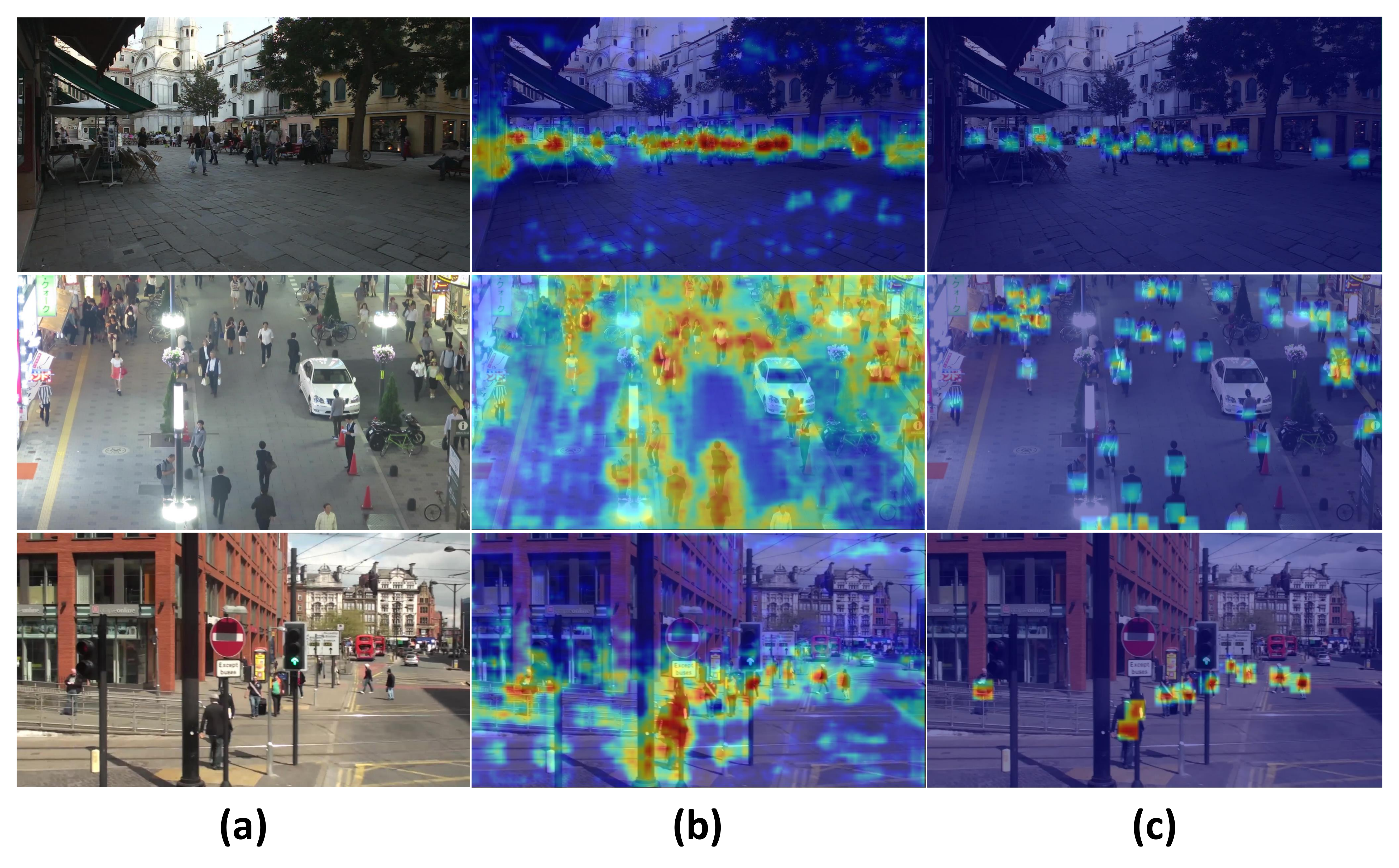}
\end{center}
\vspace{-1.2em}
\caption{Visualization of the response maps with (c) and without (b) shrinking on MOT17 dataset. To make it clear, we show the corresponding original image in (a).}
\vspace{-1em}
\label{fig:mstf}
\end{figure}

\begin{table*}[!t]
\caption{Comparison with the state-of-the-art online MOT systems under private detection protocol. We report the corresponding official metrics. ↑/↓ indicate that higher/lower is better, respectively. For a fair comparison, we obtain FPS of each method under the same experimental conditions. The best scores of methods are marked in \textcolor{red}{\bf{red}}.}

\vspace{-0.5em}
\begin{center}
\resizebox{15cm}{4.3cm}{
\begin{tabular}{ccccccccc} \toprule
Method & Published  & MOTA$\uparrow$  &  IDF1$\uparrow$  &  MT$\uparrow$  &  ML$\downarrow$   &  FP$\downarrow$  &  FN$\downarrow$  &  FPS$\uparrow$ \\ \hline

\midrule  
\multicolumn{9}{c}{\bf{MOT16}}\\
\midrule  
POI~\cite{yu2016poi}       & ECCV16 & 66.1 & 65.1 & 34.0 & 21.3 & \textcolor{red}{\bf{5061}} & 55914 & $\textless$5.2 \\
DeepSORT-2~\cite{wojke2017simple} & ICIP17 & 61.4 & 62.2 & 32.8 & 18.2 & 12852 & 56668 & $\textless$6.7 \\
HOGM~\cite{zhou2018online}     & ICPR18 & 64.8 & 73.5 & 40.6 & 22.0 & 13470 & 49927 & $\textless$8.0 \\
RAR16wVGG~\cite{fang2018recurrent} & WACV18 & 63.0 & 63.8 & 39.9 & 22.1 & 13663 & 53248 & $\textless$1.5 \\
TubeTK~\cite{pang2020tubetk} & CVPR20 & 64.0 & 59.4 & 33.5 & 19.4 & 11544 & 47502 & 1.0  \\
CTracker~\cite{peng2020chained}     & ECCV20 & 67.6 & 57.2 & 32.9 & 23.1 & 8934 & 48305 & 6.8  \\
QDTrack~\cite{QDTrack}      & CVPR21 & 69.8 & 67.1 & 41.7 & 19.8 & 9861 & 44050 & 14$\sim$30 \\
TraDeS~\cite{TraDeS}        & CVPR21 & 70.1 & 64.7 & 37.3 & 20.0 & 8091 & 45210 & 15 \\
FairMOT~\cite{zhang2020fairmot} & IJCV21 & 74.9 & 72.8 & 44.7 & 15.9 & 10163 & 34484 & \textcolor{red}{\bf{18.9}} \\
JDE~\cite{wang2019towards}       & ECCV20 & 64.4 & 55.8 & 35.4 & 20.0 & 10642 & 52523 & 18.5 \\
CSTrack~\cite{liang2020rethinking}     & Arxiv20 & 70.7 & 71.8 & 38.2 & 17.8 & 10286 & 41974 & 15.8 \\
\bf{OMC}                       &    Ours    & \textcolor{red}{\bf{76.4}} & \textcolor{red}{\bf{74.1}} & \textcolor{red}{\bf{46.1}} & \textcolor{red}{\bf{13.3}} & 10821 & \textcolor{red}{\bf{31044}} & 12.8  \\
\midrule  
\multicolumn{9}{c}{\bf{MOT17}}\\
\midrule  
TubeTK~\cite{pang2020tubetk}      & CVPR20 & 63.0 & 58.6 & 31.2 & 19.9 & 27060 & 177483 & 3.0 \\
CTracker~\cite{peng2020chained}   & ECCV20 & 66.6 & 57.4 & 32.2 & 24.2 & 22284 & 160491 & 6.8 \\
CenterTrack~\cite{zhou2020tracking}& ECCV20 & 67.8 & 64.7 & 34.6 & 24.6 & \textcolor{red}{\bf{18498}} & 160332 & \textcolor{red}{\bf{22.0}}\\
QDTrack~\cite{QDTrack}       & CVPR21 & 68.7 & 66.3 & 40.6 & 21.8 & 26589 & 146643 & 14$\sim$30 \\
TraDeS~\cite{TraDeS}       & CVPR21 & 69.1 & 63.9 & 36.4 & 21.5 & 20892 & 150060 & 15 \\
FairMOT~\cite{zhang2020fairmot}    & IJCV21 & 73.7 & 72.3 & 43.2 & 17.3 & 27507 & 117477 & 18.9  \\
CSTrack~\cite{liang2020rethinking}    & Arxiv20 & 70.6 & 71.6 & 37.5 & 18.7 & 24804 & 137832 & 15.8 \\
\bf{OMC}                          &    Ours    & \textcolor{red}{\bf{76.3}} & \textcolor{red}{\bf{73.8}} & \textcolor{red}{\bf{44.7}} & \textcolor{red}{\bf{13.6}} & 	28894 & \textcolor{red}{\bf{101022}} & 12.8  \\
\midrule  %
\multicolumn{9}{c}{\bf{MOT20}}\\
\midrule  
FairMOT~\cite{zhang2020fairmot}    & IJCV21 & 61.8 & 67.3 & \textcolor{red}{\bf{68.8}} & \textcolor{red}{\bf{7.6}} & 103440 & \textcolor{red}{\bf{88901}} & \textcolor{red}{\bf{8.4}}  \\
\bf{OMC }                           &    Ours    & \textcolor{red}{\bf{70.7}} & \textcolor{red}{\bf{67.8}} & 56.6 & 13.3 & \textcolor{red}{\bf{22689}} & 125039 & 6.7  \\
\bottomrule
\end{tabular}}
\vspace{-1.5em}
\end{center}
\label{tab:comparison}%
\end{table*}

\noindent\textbf{Understanding the Re-check Network.}
To understand the impact of the re-check network, we evaluate the tracker (Baseline-BAR) with different variants of the re-check network, as shown in Tab.~\ref{tab:recheck_a}. Firstly, we replace the global search with the standard local search, \ie, considering the neighborhood region of previous targets in the last frame~\cite{dasot}. Comparing the results of \ding{173} and \ding{177}, we find that with a global view, our tracker can more accurately propagate previous tracklets to current frame, which achieves better tracking performance on FP, FN, MOTA and IDF1 scores. Secondly, we discard the shrinking operation in Eq.~\ref{eq:shrink}. As the result shown in \ding{174}, without shrinking, the FP number dramatically increases, which eventually causes the decrease of MOTA score. We visualize the shrinking and non-shrinking response maps in Fig.~\ref{fig:mstf}, which shows that the shrinking operation can filter most false-positive responses and effectively keep the transductions of previous targets. Furthermore, we conduct two ablation experiments to prove the rationality of the refinement module design, as shown in \ding{175} and \ding{176}. During calculating the similarity map $\mathbf{M}_p$, we involve visual feature $\mathbf{F}_t$ to mitigate false-positive transduction. The result of \ding{175} verifies that introducing $\bm{F}_t$ can effectively decrease FP number, \ie, 67885 $\rightarrow$ 28894. When we discard the inverted bottleneck module (IBM), the MOTA score decreases from 76.3 to 75.7 (\ding{176} \emph{vs}. \ding{177}). It confirms that the precoding of similarity map and the semantic information in visual feature can complement each other for better tracking performance.

\subsection{Comparison on MOT Benchmarks}
\begin{figure}[t]
\begin{center}
\includegraphics[width=1\linewidth]{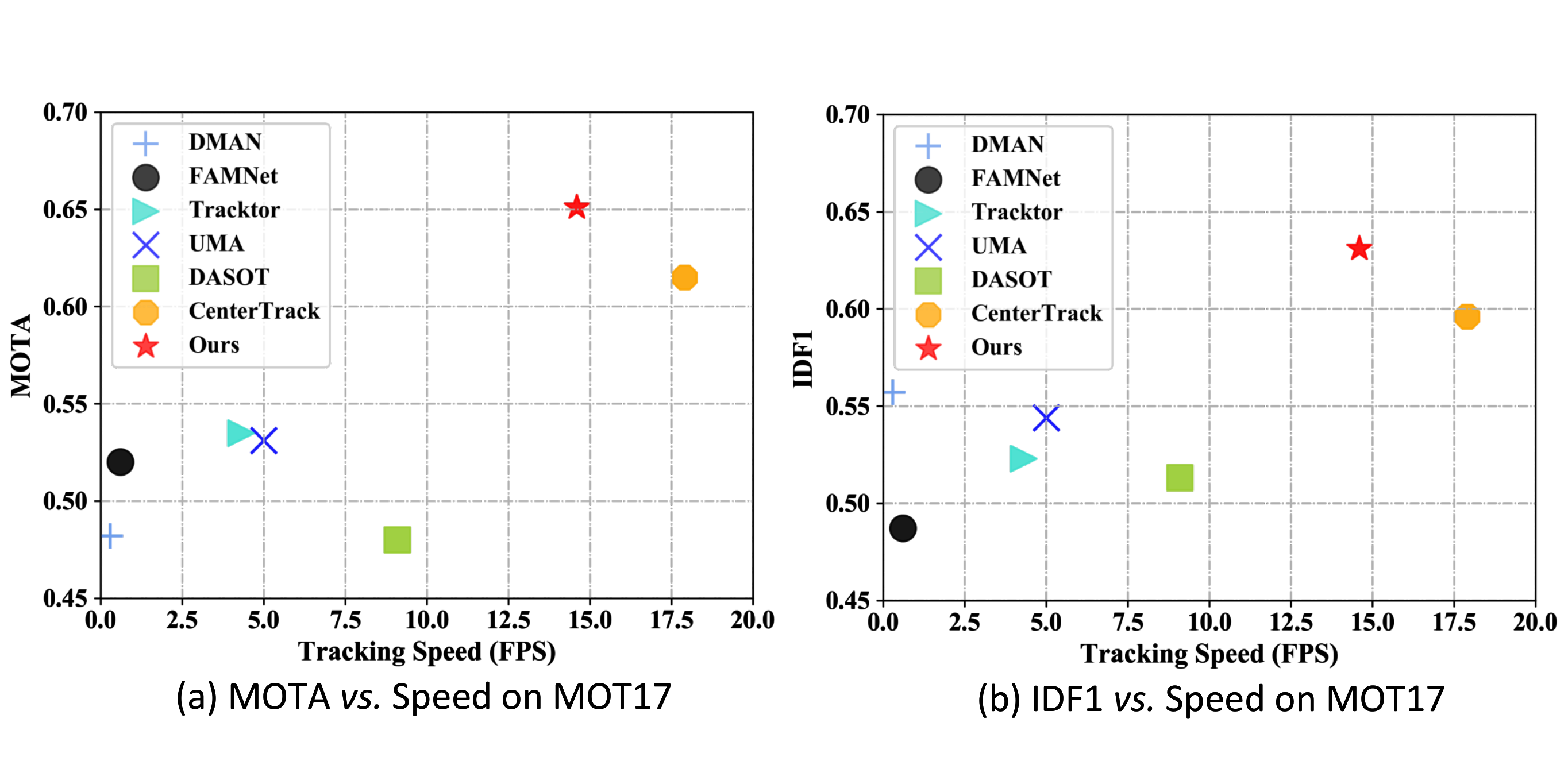}
\end{center}
\vspace{-1em}
\caption{Tracking performance (MOTA and IDF1) and tracking speed (FPS) of the proposed method and other MOT methods on MOT17 benchmark.}
\label{fig:compare}
\vspace{-1em}
\end{figure}

\noindent\textbf{State-of-the-art Comparison.}
We compare OMC with other state-of-the-art MOT methods on the testing sets of MOT16, MOT17 and MOT20. For evaluating on the MOT20 benchmark, we fine-tune it on the training set of MOT20 following the same training procedure. As shown in Tab. ~\ref{tab:comparison}, our tracker achieves new state-of-the-art MOTA and IDF1 scores on all three benchmarks. Specifically, the proposed OMC outperforms the recent state-of-the-art tracker FairMOT~\cite{zhang2020fairmot} by $1.5\sim8.9$ points on MOTA. 

\noindent\textbf{Methods with Temporal Cues Mining.} To better illustrate the effectiveness of the proposed method, we compare our tracker with the advanced MOT methods using temporal cues under the same public detection protocol~\cite{milan2016mot16}. The compared methods are divided into two categories, \ie, one using SOT trackers which include DMAN~\cite{zhu2018online}, FAMNet~\cite{chu2019famnet}, UMA~\cite{yin2020unified} and DASOT~\cite{dasot}, and the other directly propagating previous tracklets by predicting the offset of bounding boxes in the last frame, \ie,  Tracktor~\cite{bergmann2019tracking} and CenterTrack~\cite{zhou2020tracking}. As shown in Fig.~\ref{fig:compare}, our method runs faster (14.6 FPS \emph{vs}. $0.3\sim9.1$ FPS) and achieves better tracking performance compared with other methods using SOT trackers. Moreover, comparing our tracker with methods directly propagating previous tracklets, our method still gains best tracking performance on MOTA and IDF1. 

\section{Conclusion}
This work has presented a novel double-check approach for MOT, to reload the ``fake background" that is caused by the detector's over-reliance on the single-frame visual cues. Unlike prior attempts, we propose a novel re-check network, which can mining temporal cues with a small overhead. Concretely, we expand the role of ID embeddings from data association to motion forecasting and propagate the previous tracklets to the current frame using global embedding search. Based on this, we construct a new one-shot MOT tracker, namely OMC, which integrates detection, embedding extraction and temporal cues mining into a unified framework. The quantitative experimental results have shown that the re-check network can restore the targets missed by the detector more effectively than the prior methods. OMC is simple, efficient, and achieves new state-of-the-art performance on MOT16, MOT17, and MOT20.

{\small
\bibliography{aaai22}
}

\clearpage

\twocolumn[
\begin{@twocolumnfalse}
	\section*{\centering{One More Check: Making ``Fake Background'' Be Tracked Again \\ ------Supplementary Material------\\[25pt]}}
\end{@twocolumnfalse}
]

In the supplementary material, we provide additional experiments and the qualitative results on MOT17~\cite{milan2016mot16} and MOT20~\cite{dendorfer2020mot20} testing set. Furthermore, an additional section (Discussion) is presented to explain some questions for readers. 

\section*{Experiments}
\begin{figure}[t]
\begin{center}
\includegraphics[width=1\linewidth]{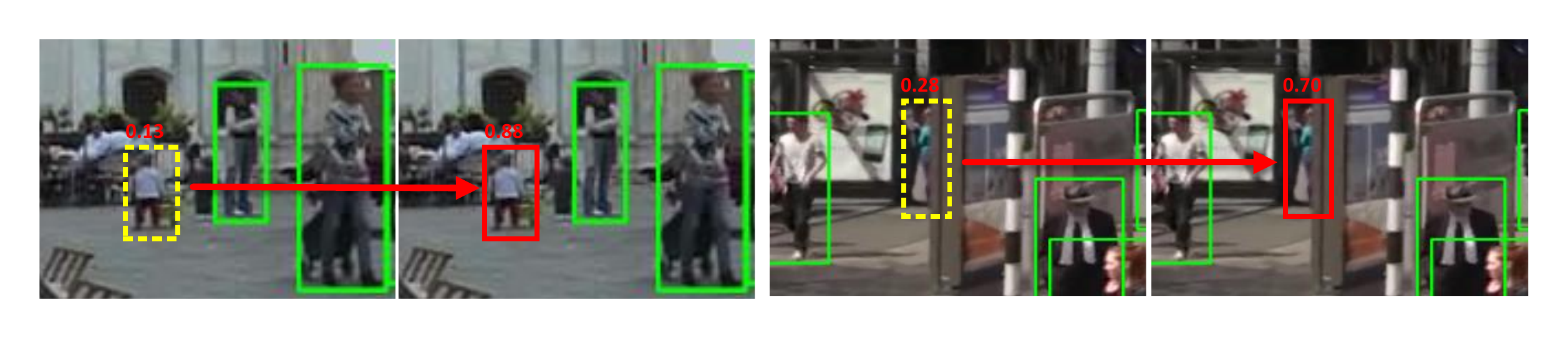}
\end{center}
\vspace{-1em}
\caption{Several examples of the restored results.}
\label{fig:restore_example}
\end{figure}

\noindent\textbf{Restored Results of Re-check Network.}
In this section, we further list the number of restored bounding boxes (confidence threshold $<$ 0.5) in Tab.~\ref{Tab:restore}. The experimental results show that the re-check network can efficiently repair the misclassified targets, even they have very low confidence. Furthermore, we visualize several samples in Fig.~\ref{fig:restore_example}.

\noindent\textbf{Tracking under Public Detection.}
For a fair comparison with other MOT methods using temporal cues, we compare all trackers on the MOT17 benchmark under the same public detection protocol~\cite{milan2016mot16}. Specifically, we replace the detection results of our tracker (\ie, $\bm{D}_{base}$) with the same official detection results. We only use public detection to initialize a new trajectory if it is not near tracked boxes in the current frame. All the other bounding boxes in our results are from the tracking boxes of the re-check network. Note that this strategy is allowed and is commonly used by the online methods using temporal cues~\cite{bergmann2019tracking,zhou2020tracking,keuper2018motion,chen2018real}. We report more metrics of quantitative tracking results in Tab.~\ref{tab:comparison}. Compared with other online trackers using temporal cues, our tracker can significantly decrease FN and achieve better tracking performance on all four metrics.

\noindent\textbf{``Plug-and-Play'' Module.}
In this section, we present the proposed re-check network is a ``plug-and-play'' module that can work well with other one-shot trackers, \eg, FairMOT~\cite{zhang2020fairmot}. We integrate the re-check network (R) into FairMOT as the same protocol as CSTrack~\cite{liang2020rethinking} without any special modifications and hyper-parameter tuning. As shown in Tab.~\ref{tab:fairmot}, the re-check network can efficiently restore the ``fake background'', decreasing FP from 117447 to 108556. Although the performance of FairMOT is high enough, the re-check network brings obvious gains (MOTA +1 point and IDF1 +1.5 points). This again proves the effectiveness and generality of re-check network.

Due to the limited submissions of the MOT Challenge (4 times per model on each benchmark), the following ablation experiments are conducted on MOT15, \ie, training on MOT17 training set and testing on MOT15 training set. For a fair comparison, the overlapped videos are discarded.

\begin{table}[t]
    \begin{center}
        \caption{Restored results on MOT17 testing set.} 
        \fontsize{8pt}{3.7mm}\selectfont
        \begin{threeparttable}
            \begin{tabular}{@{}c@{} | @{}c@{} @{}c@{} @{}c@{} @{}c@{} @{}c@{} | @{}c@{}}
                \cline{1-7}
                 Conf. &~~0.4$\sim$0.5~~&~~0.3$\sim$0.4~~&~0.2$\sim$0.3~~&~~0.1$\sim$0.2~~&~~$<$0.1~~&~~All Restored~~
                \\ 
                \cline{1-7} 
                ~~Number~~&~~6093& 7794 & 6794 & 9855 &~~2439~~ & 32975 \\
				Ratio &~~18.5\% &~~23.6\% & 20.6\% & 29.9\% & 7.4\% & 100\%  \\     
                \cline{1-7}
            \end{tabular}

        \end{threeparttable}
        \label{Tab:restore}
    \end{center}
\end{table}

\begin{table*}[t]
\caption{Comparison with the state-of-the-art online MOT systems using temporal cues under public detection on MOT17 benchmark~\cite{milan2016mot16}. We report the corresponding official metrics. ↑ indicates that higher is better, ↓ indicates that lower is better. The best scores of methods are marked in \textcolor{red}{\bf{red}}.}
\begin{center}
\resizebox{16cm}{1.6cm}{
\begin{tabular}{ccccccccc} \toprule
Method & Published  & MOTA$\uparrow$  &  IDF1$\uparrow$  &  MT$\uparrow$  &  ML$\downarrow$   &  FP$\downarrow$  &  FN$\downarrow$  &  FPS$\uparrow$ \\ \hline
DMAN~\cite{zhu2018online} & ECCV18 & 48.2 & 55.7 & 19.3 & 38.3 & 26218 & 263608 & 0.3 \\
FAMNet~\cite{chu2019famnet} & ICCV19 & 52.0 & 48.7 & 19.1 & 33.4 &  14138 & 253616 & 0.6 \\
Tracktor++~\cite{bergmann2019tracking} & ICCV19 & 53.5 & 52.3 & 19.5 & 36.6 & \textcolor{red}{\bf{12201}} & 248047 & $\textless$5.0 \\
DASOT~\cite{dasot} & AAAI20 & 48.0 & 51.3 & 19.9 & 34.9 & 38830 & 250533 & 9.1 \\
UMA~\cite{yin2020unified} & CVPR20 & 53.1 & 54.4 & 21.5 & 31.8 & 22893 & 239534 & 5.0 \\
CenterTrack~\cite{zhou2020tracking}& ECCV20 & 61.5 & 59.6 & 26.4 & 31.9 & 14076 & 200672 & \textcolor{red}{\bf{17.5}} \\
\bf{OMC (Ours)}                &    Ours   & \textcolor{red}{\bf{65.1}} & \textcolor{red}{\bf{63.1}} & \textcolor{red}{\bf{27.4}} & \textcolor{red}{\bf{23.2}} & 22320 & \textcolor{red}{\bf{179772}} & 14.6  \\
\bottomrule
\end{tabular}}

\end{center}
\label{tab:comparison}%
\end{table*}

\begin{table}[t]
    \begin{center}
        \caption{Results of FairMOT with re-check network (R) on MOT17.} 
        \fontsize{8pt}{4.2mm}\selectfont
        \begin{threeparttable}
            \begin{tabular}{@{}c@{} | @{}c@{} @{}c@{}  @{}c@{}  @{}c@{} @{}c@{} @{}c@{}}
                \cline{1-7}
                ~~Method~~&~~MOTA$\uparrow$~~&~~IDF1$\uparrow$~~&~~MT$\uparrow$~~&~~ML$\downarrow$~~&~~~~~FP$\downarrow$~~~&~~~~~FN$\downarrow$~~~
                \\
                \cline{1-7} 
                FairMOT & 73.7 & 72.3 & 43.2 & 17.3 & \bf{27507} & 117477 \\
				~~FairMOT+R~~  & \bf{74.7} & \bf{73.8} & \bf{44.3} & \bf{15.4} & 30162 & \bf{108556}\\
                \cline{1-7}
            \end{tabular}

        \end{threeparttable}
        \label{tab:fairmot}
    \end{center}
\end{table}

\noindent\textbf{Impact of Response Region Shrinking.}
We shrink the high response region after attaining similarity map $\bm{M}$, as in Eq.~\textcolor{red}{5}. In Tab.~\ref{tab:mstf}, we present the influence of different shrinking radius. Compared \ding{177} with \ding{172}$\sim$\ding{176}, without shrinking, the false positives (FP) would dramatically increase, which eventually causes the decrease of the MOTA score. We set the radius to 3, as it provides the best MOTA score.

\begin{figure*}[t]
\begin{center}
\includegraphics[width=0.85\linewidth]{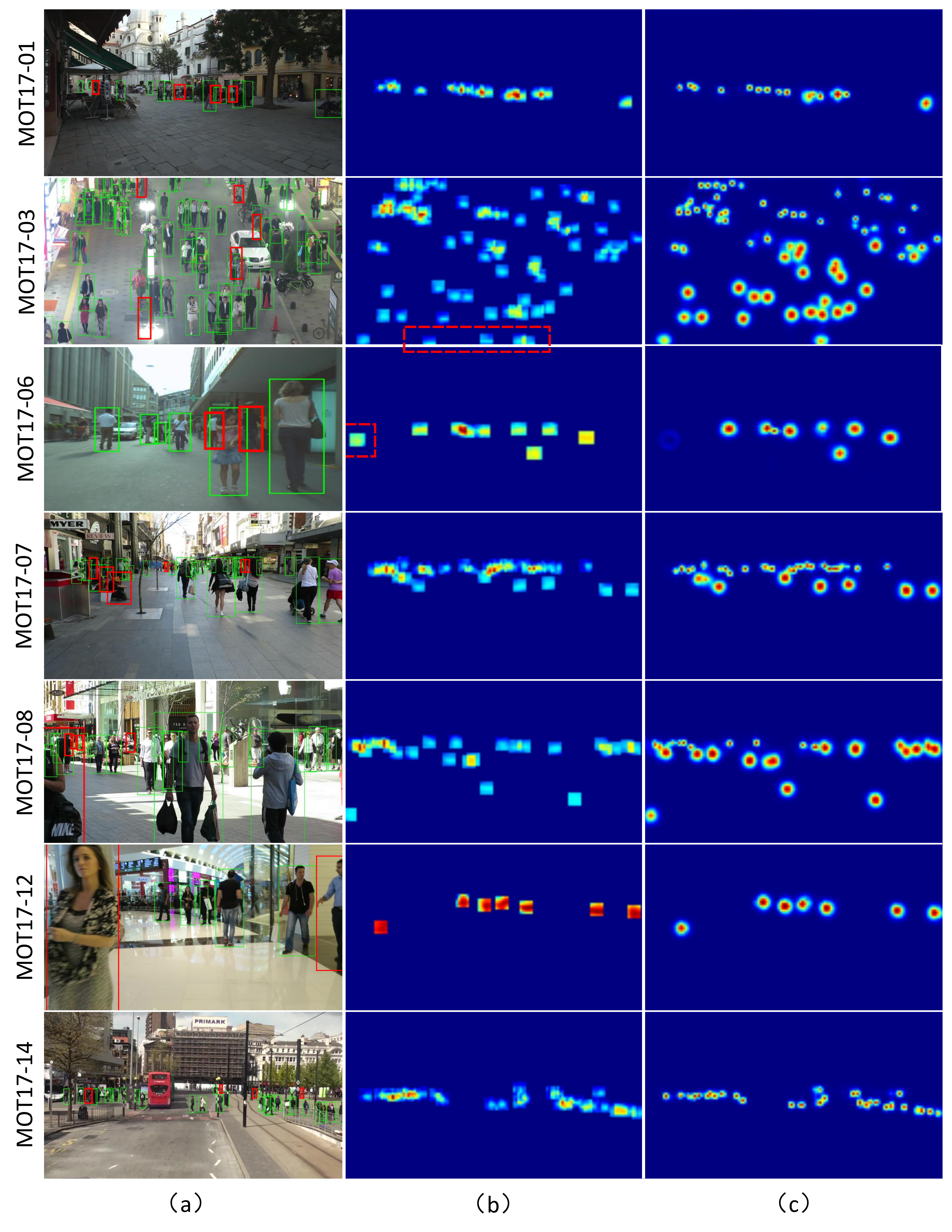}
\end{center}
\vspace{-1em}
\caption{Visualization about transductive results of re-check network on MOT17 testing set: (a) final candidate bounding boxes, where the green boxes indicate basic detections and red boxes indicate auxiliary boxes from transductive detections. (b) the aggregated similarity map $\bm{M}_s$. (c) the final prediction from re-check network $\bm{M}_p$. Best viewed in color and zoom in. }
\label{fig:vis}
\end{figure*}

\begin{table}[!t]
    \begin{center}

        \caption{Impact of shrinking high response region.}
        \fontsize{9pt}{4.3mm}\selectfont
        \begin{threeparttable}
            \begin{tabular}{ @{}c@{} | @{}c@{} | @{}c@{}  @{}c@{} @{}c@{}  @{}c@{} @{}c@{} @{}c@{}}
                \cline{1-8}
                \#NUM ~& $r$ & ~MOTA$\uparrow$~  & ~IDF1$\uparrow$~  & ~~MT$\uparrow$~~ &  ~ML$\downarrow$~ & ~FP$\downarrow$~ &  ~FN$\downarrow$~
                \\
                \cline{1-8} 
                 \ding{172} & 1 & 77.3 & \bf{73.8} & \bf{75.3} & 11.7 & ~2395~ & ~1480~\\
                \ding{173} & 3 & \bf{77.6} & 73.6 & 73.9 & 12.0 & \bf{2288} & 1527\\
                \ding{174} & 5 & 77.3 & 73.4 & 73.9 & 12.0 & 2331 &1547\\
                \ding{175} & 7 & 77.0 & 73.4 & 73.9 & 12.0 & 2424 &1501\\
                \ding{176} & 9 & 76.4 & 73.4 & 75.1 & 11.7 & 2543 &1483\\
                \ding{177} & ~\emph{w/o} shrink~ & 75.9 & \bf{73.8} & 74.8 & \bf{10.9} & 2651 &\bf{1470}\\
                \cline{1-8}

            \end{tabular}
        \end{threeparttable}
         \label{tab:mstf}
    \end{center}
    \vspace{-0.15cm}
\end{table}

\noindent\textbf{Influence of Fusion Threshold.}
For fusing the transductive detections $\bm{D}_{trans}$ and basic detections $\bm{D}_{base}$, we calculate the targetness score $s$ for each bounding box $\bm{b}_i$ in $\bm{D}_{trans}$, and retrain the score above threshold $\epsilon$ as complement of basic detections $\bm{D}_{base}$. To study the impact of fusion threshold $\epsilon$ in our tracker, we perform an ablation experiment on MOT15~\cite{leal2015motchallenge}, as shown in Tab.~\ref{tab:ft}. \ding{175}$\sim$\ding{181} demonstrate the robustness of our tracker with respect to fusion threshold $\epsilon$, when $\epsilon \in [0.4,1]$. Comparing \ding{172}$\sim$\ding{174} with \ding{175}$\sim$\ding{181}, we observe that low fusion thresholds degrade tracking performance. The possible reason is that some targets would be assigned with two boxes and one of them will be considered as false positive. We set $\epsilon$ to 0.5 considering its best MOTA and IDF1 score.

\begin{table}[!t]
    \begin{center}

        \caption{Impact of fusion threshold $\epsilon$ in our tracker.}
        \fontsize{9pt}{4.3mm}\selectfont
        \begin{threeparttable}
            \begin{tabular}{ @{}c@{} | @{}c@{} | @{}c@{}  @{}c@{} @{}c@{}  @{}c@{} @{}c@{} @{}c@{}}
                \cline{1-8}
                \#NUM ~& ~~~~~~$\epsilon$~~~~~~ & ~MOTA$\uparrow$~  & ~IDF1$\uparrow$~  & ~~MT$\uparrow$~~ &  ~ML$\downarrow$~ & ~~FP$\downarrow$~~ &  ~~FN$\downarrow$~~
                \\
                \cline{1-8} 
                \ding{172} & 0.1 & 47.4 & 59.4 & 74.8 & \bf{11.1} & 6787 & 1746\\
                \ding{173} & 0.2 & 72.1 & 70.3 & \bf{75.1} & 11.4 & 3116 & 1525\\
                \ding{174} & 0.3 & 76.6 & \bf{73.8} & 74.8 & 11.7 & 2454 & \bf{1496}\\
                \ding{175} & 0.4 & 77.4 & 73.5 & 73.9 & 12.0 & 2330 & 1506\\
                \ding{176} & 0.5 & \bf{77.6} & 73.6 & 73.9 & 12.3 & 2289 & 1526\\
                \ding{177} & 0.6 & \bf{77.6} & 73.4 & 73.9 & 12.0 & 2284 & 1542\\
                \ding{178} & 0.7 & 77.5 & 73.5 & 73.9 & 12.3 & 2276 & 1565\\
                \ding{179} & 0.8 & 77.3 & 73.1 & 73.0 & 12.3 & 2261 & 1605\\
                \ding{180} & 0.9 & 77.3 & 73.1 & 72.7 & 13.2 & 2223 & 1643\\
                \ding{181} & 1.0 & \bf{77.6} & 73.2 & 71.8 & 13.2 & \bf{2161} & 1645\\
                \cline{1-8}

            \end{tabular}
        \end{threeparttable}
         \label{tab:ft}
    \end{center}
    \vspace{-0.15cm}
\end{table}

\vspace{1mm}
\noindent\textbf{Impact of ID Embedding Generation.}
During tracking, ID embeddings of previous tracklets are linearly updated~\cite{liang2020rethinking}. Alternatively, we can simply use embeddings of their first or last occurrence as the temporal cues. We compare their influence on tracking performance in Tab.~\ref{tab:idchoose}. Comparing \ding{172} with \ding{174}, we obverse that when using ID embedding of the first appearance, the false positives (FP) would dramatically increase. In contrast, using the ID embedding from the last frame achieves better tracking performance, \ie, 77.4 MOTA score, but it is still inferior to that of using the updated ID embedding (\ding{173} \emph{vs}. \ding{174}).

\begin{table}[!t]
    \begin{center}

        \caption{Impact of ID embedding generation in our tracker.}
        \fontsize{9pt}{4.3mm}\selectfont
        \begin{threeparttable}
            \begin{tabular}{ @{}c@{} | @{}c@{} | @{}c@{}  @{}c@{} @{}c@{}  @{}c@{} @{}c@{} @{}c@{}}
                \cline{1-8}
                \#NUM ~&~Generation~& ~MOTA$\uparrow$~  & ~IDF1$\uparrow$~  & ~~MT$\uparrow$~~ &  ~ML$\downarrow$~ & ~~FP$\downarrow$~~ &  ~~FN$\downarrow$~~
                \\
                \cline{1-8} 
                \ding{172} & First & 76.4 & 73.1 & 75.1 & 12.0 & 2501 & 1529\\
                \ding{173} & Last & 77.4 & 73.4 & \bf{76.2} & \bf{11.4} & 2377 & \bf{1492}\\
                \ding{174} & Updated & \bf{77.6} & \bf{73.6} & 73.9 & 12.3 & \bf{2289} & 1526\\
                \cline{1-8}

            \end{tabular}
        \end{threeparttable}
         \label{tab:idchoose}
    \end{center}
    \vspace{-0.15cm}
\end{table}

\noindent\textbf{Visualization of Transductive Results.}
We present the aggregated similarity map $\bm{M}_s$ and the final prediction of re-check network $\bm{M}_p$ in Fig.~\ref{fig:vis} (b) and (c), respectively. The former indicates that our method is effective in transferring previous tracklets to the current frame. The latter represents the proposed refinement module in the re-check network can effectively filter false positives in the similarity map $\bm{M}_s$. Concretely, in Fig.~\ref{fig:vis} (a), the \textcolor{red}{red} boxes denote ones missed by the basic detections and restored by the transductive detections. It exhibits the efficacy of the proposed re-check network on restoring ``fake background'' and making the missed targets be tracked again. As illustrated by the MOT17-03 and MOT17-06 sequence, the false positives (\textcolor{red}{in the red dotted box}) are successfully filtered by the refinement network. Moreover, our method can perform robust tracking under occlusion scenes (see MOT17-07 and MOT17-08) and facing small targets (see MOT17-14).

\begin{figure*}[t]
\begin{center}
\includegraphics[width=0.93\linewidth]{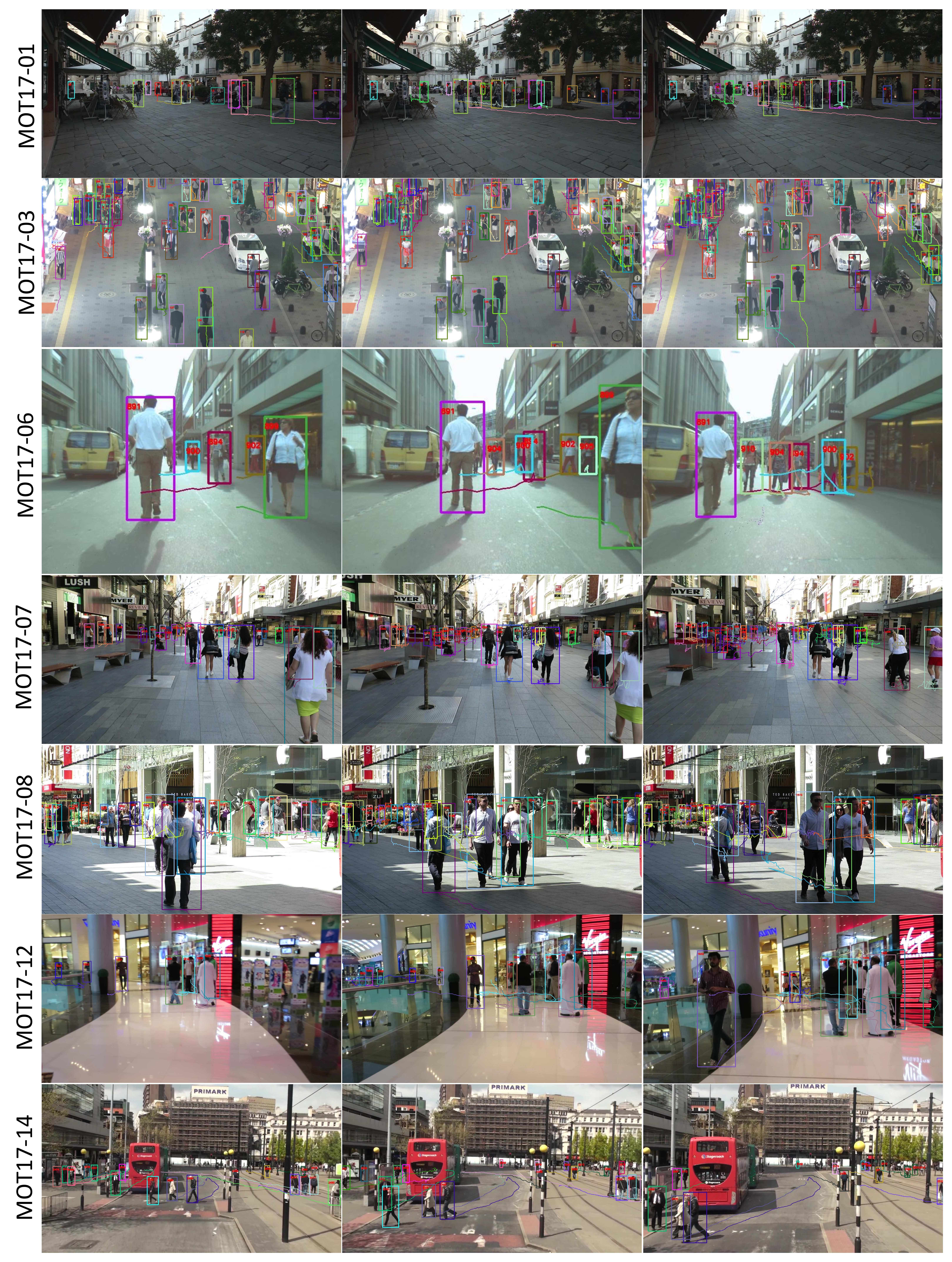}
\end{center}
\caption{The qualitative results of our method on sequences from MOT17 test set. Bounding boxes, identities and trajectories are marked in the images, where bounding boxes with different colors represent different targets. Best viewed in color.}
\label{fig:17}
\end{figure*}

\begin{figure*}[t]
\begin{center}
\includegraphics[width=0.93\linewidth]{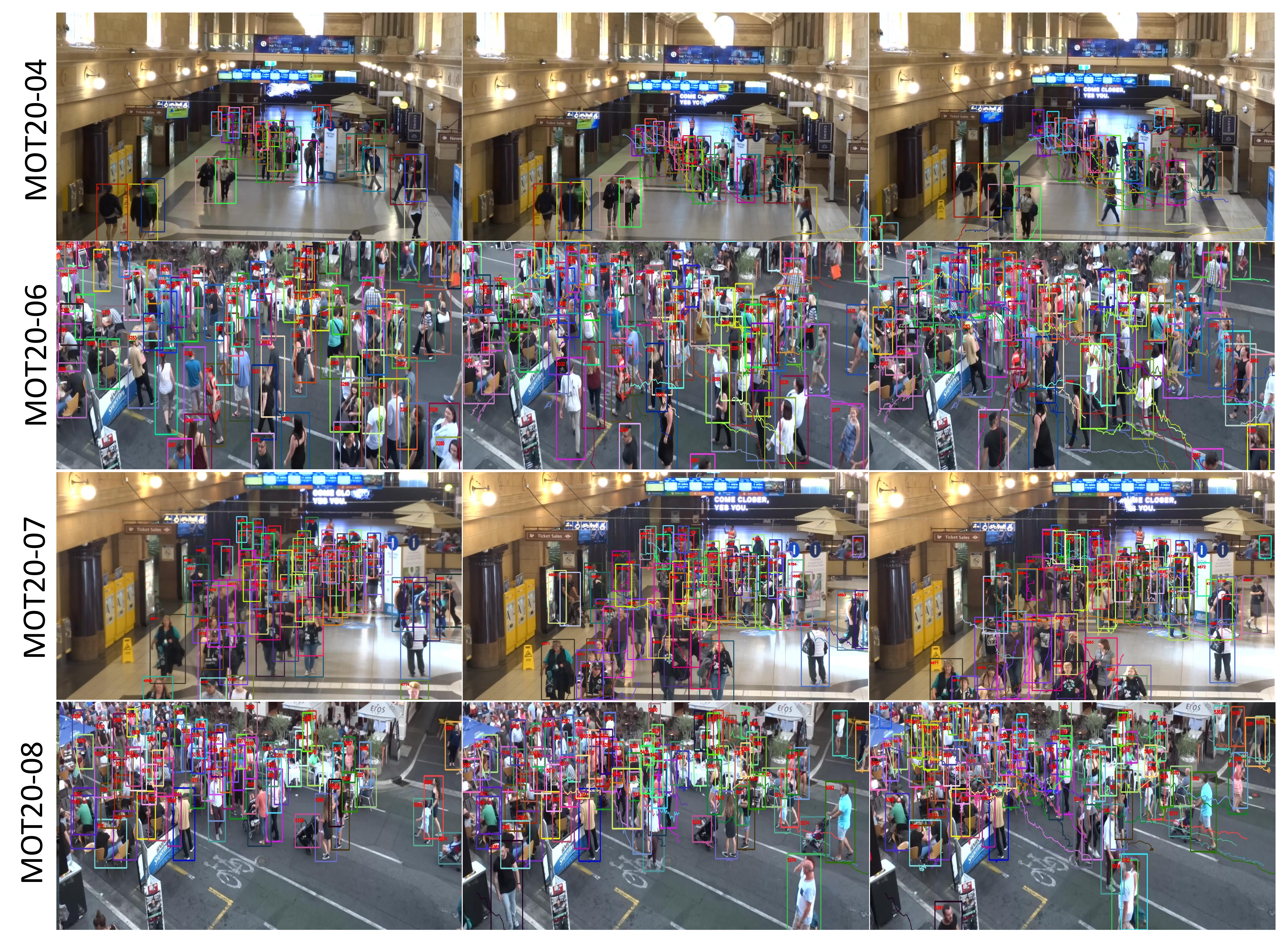}
\end{center}
\caption{The qualitative results of our method on sequences from MOT20 test set. Bounding boxes, identities and trajectories are marked in the images, where bounding boxes with different colors represent different targets. Best viewed in color.}
\label{fig:20}
\end{figure*}

\vspace{1mm}
\noindent\textbf{Qualitative Tracking Results.}
In this section, we visualize qualitative tracking results of each sequence in MOT17 and MOT20 test set, as shown in Fig.~\ref{fig:17} and Fig.~\ref{fig:20}. From the results of MOT17-03, we can see that our method performs well at the boundary.  The results of MOT17-07 show that our method can detect small objects accurately. Note, our tracker can perform robust tracking under occlusion scenes (see MOT20), where targets are heavily occluded.

\section*{Discussion}
\noindent\textcolor{blue}{\textbf{Q1:} \emph{When the targets are not detected for several frames, how does the re-check network deal with those targets?}}

\noindent\textbf{R1:} During tracking inference, a tracklet will be reserved unless there is no matched bounding box for K successive frames (K=30, following CSTrack). The re-check network uses ID embeddings of all previous tracklets for propagation rather than that only in the last frame. In other words, all targets appearing in the previous 30 frames are considered for tracklets propagation. Therefore, even the targets are not detected for several frames, the re-check network can still handle them by using their appearance feature in the previous correctly detection frames. 

\begin{table}[t]  
    \begin{center}
        \caption{Hard mask \emph{vs.} soft mask.}
        \fontsize{8.5pt}{4mm}\selectfont
        \begin{threeparttable}
            \begin{tabular}{ @{}c@{} | @{}c@{} | @{}c@{}  @{}c@{} @{}c@{}  @{}c@{} @{}c@{} @{}c@{}}
                \cline{1-8}
                \#NUM ~&  & ~MOTA$\uparrow$~  & ~IDF1$\uparrow$~  & ~~MT$\uparrow$~~ &  ~ML$\downarrow$~~&~~FP$\downarrow$~~~&~~~FN$\downarrow$~
                \\
                \cline{1-8} 
                \ding{172} & ~ hard mask~ & \bf{76.3} & \bf{73.8} & 44.7 & 13.6 &~\bf{28894}~&~\bf{101022}~~\\
                \ding{173} &  ~soft mask~ & 76.0 & 73.0 & \bf{45.1} & \bf{13.2} & 29993 & 101447\\
                \cline{1-8}

            \end{tabular}
        \end{threeparttable}
         \label{Tab:softmask}
    \end{center}
    \vspace{-1.5em}
\end{table}

\noindent\textcolor{blue}{\textbf{Q2:} \emph{whether the soft mask will improve the performance of re-check network?}}

\noindent\textbf{R2:} We try to replace the hard mask with a Gaussian-like soft mask to shrink the scope of high responses. As shown in Tab.~\ref{Tab:softmask}, it achieves comparable performance with the hard mask version. However, considering the MOTA score, the release version still uses hard mask.

\end{document}